\documentclass[10pt]{article} 

\usepackage[accepted]{styles/rlc}
\usepackage{hyperref}
\usepackage{microtype}
\usepackage{graphicx}
\usepackage{subfigure}
\usepackage{booktabs}

\usepackage{amsmath}
\usepackage{amssymb}
\usepackage{mathtools}
\usepackage{amsthm}
\usepackage{scalerel}
\usepackage[shortlabels]{enumitem}

\usepackage[capitalize,noabbrev,nameinlink]{cleveref}
\creflabelformat{equation}{#2#1#3}
\usepackage{csquotes}
\usepackage{multirow}
\usepackage{mathrsfs}
\usepackage{mdframed}
\usepackage{bm}
\usepackage{tipa}
\usepackage{mathbbol}
\DeclareSymbolFontAlphabet{\mathbb}{AMSb}
\DeclareSymbolFontAlphabet{\mathbbl}{bbold}

\usepackage{xcolor}
\definecolor{dred}{RGB}{153,80,43}
\definecolor{dblue}{RGB}{0,114,178}
\hypersetup{
    colorlinks=true,
    breaklinks=true,
    urlcolor=dblue, 
    linkcolor=dred, 
    citecolor=dblue
}
\usepackage{caption}

\def\environment{e}
\def\env{\environment}

\newcommand\reaches{%
  \mathrel{\ooalign{\hss$\rightsquigarrow$\hss\kern-1.45ex\raise1.0ex\hbox{{\scaleobj{0.55}{\env}}}\hspace{4pt}}}}
\newcommand\cnot[1]{%
  \mathrel{\ooalign{\hfil$#1$\hfil\cr\hfil$/$\hfil\cr}}}
\newcommand\notreaches{%
  \mathrel{\ooalign{\hss$\cnot\rightsquigarrow$\hss\kern-1.45ex\raise1.0ex\hbox{{\scaleobj{0.55}{\env}}}\hspace{4pt}}}}

\newcommand\alwaysreach{%
  \mathrel{\ooalign{\hss$\square$\hss\kern-0.22ex\hbox{{$\reaches$}}}}}

\newcommand\generates{%
 \mathrel{\ooalign{\hss$\vdash$\hss\kern-0.65ex\raise0.9ex\hbox{{\scaleobj{0.55}{\env}}}}}}

\usepackage{calligra}
\DeclareMathAlphabet{\mathcalligra}{T1}{calligra}{m}{n}

\newtheorem*{theorem*}{Theorem}

\newtheorem*{definition*}{Definition}

\numberwithin{equation}{section}
\numberwithin{theorem}{section}
\numberwithin{definition}{section}
\numberwithin{conjecture}{section}

\usepackage{etoolbox}
\makeatletter
\patchcmd{\NAT@test}{\else \NAT@nm}{\else \NAT@hyper@{\NAT@nm}}{}{}
\makeatother

\usepackage{tikz}

\usepackage[textsize=tiny]{todonotes}

\newcommand{\creflink}[1]{\hyperref[#1]{\textcolor{blue}{\cref{#1}}}}

\usepackage{newpxmath,newpxtext}

\title{Three Dogmas of Reinforcement Learning}
	
\author{David Abel\\
dmabel@google.com\\
Google DeepMind
\And
Mark K. Ho \\
mkh260@nyu.edu \\
New York University
\And
Anna Harutyunyan \\
harutyunyan@google.com \\
Google DeepMind}

\begin{document}
\maketitle

\begin{abstract}
Modern reinforcement learning has been conditioned by at least three dogmas.
%
The first is the \textit{environment spotlight}, which refers to our tendency to focus on modeling environments rather than agents.
%
The second is our treatment of \textit{learning as finding the solution to a task}, rather than adaptation.
%
The third is the \textit{reward hypothesis}, which states that all goals and purposes can be well thought of as maximization of a reward signal.
%
These three dogmas shape much of what we think of as the science of reinforcement learning. While each of the dogmas have played an important role in developing the field, it is time we bring them to the surface and reflect on whether they belong as basic ingredients of our scientific paradigm.
In order to realize the potential of reinforcement learning as a canonical frame for researching intelligent agents, we suggest that it is time we shed dogmas one and two entirely, and embrace a nuanced approach to the third.
\end{abstract}

\section{On a Paradigm for Intelligent Agents}

In \textit{The Structure of Scientific Revolution}, Thomas Kuhn distinguishes between two phases of scientific activity \citep{kuhn2012structure}. The first Kuhn calls "normal science" which he likens to puzzle-solving, and the second he calls the "revolutionary" phase, which consists of a re-imagining of the basic values, methods, and commitments of the science that Kuhn collectively calls a "paradigm".

The history of artificial intelligence (AI) arguably includes several swings between these two phases, and several paradigms. The first phase began with the 1956 Dartmouth workshop \citep{mccarthy2006proposal} and arguably continued up until sometime around the publication of the report by \citet{james1973artificial} that is thought to have heavily contributed to the onset of the first AI winter \citep{haenlein2019brief}. In the decades since, we have witnessed the rise of a variety of methods and research frames such as symbolic AI \citep{newell1961gps,newell2007computer}, knowledge-based systems \citep{buchanan1969heuristic} and statistical learning theory \citep{vapnik1971uniform,Valiant1984,cortes1995support}, culminating in the most recent emergence of deep learning \citep{krizhevsky2012imagenet,lecun2015deep,vaswani2017attention} and large language models \citep{brown2020language,bommasani2021opportunities,achiam2023gpt}. 

In the last few years, the proliferation of AI systems and applications has hopelessly outpaced our best scientific theories of learning and intelligence. Yet, it is our duty as scientists to provide the means to understand the current and future artifacts borne from the field, especially as these artifacts are set to transform society. It is our view that reflecting on the current paradigm and looking beyond it is a key requirement for unlocking this understanding.

In this position paper, we make two claims. First, reinforcement learning (RL) is a good candidate for a complete paradigm for the science of intelligent agents, precisely because "it explicitly considers the whole problem of a goal-directed agent interacting with an uncertain environment" (p. 3, \citeauthor{sutton2018reinforcement},  \citeyear{sutton2018reinforcement}).
Second, in order for RL to play this role, we must reflect on the ingredients of our science and shift a few points of emphasis. These shifts are each subtle departures from three "dogmas", or implicit assumptions, summarized as follows:
\begin{enumerate}
    \item \textsc{The Environment-Spotlight} (\cref{sec:dogma1_env_spotlight}): Our emphasis on modeling environments rather than agents.
    
    \item \textsc{Learning as Finding a Solution} (\cref{sec:dogma2_learning_as_solving}): Our search for agents that learn to solve tasks.
    
    \item \textsc{The Reward Hypothesis} (\cref{sec:dogma3_rh}): Assuming all goals are well thought of in terms of reward maximization.
\end{enumerate}

When we relax these dogmas, we arrive at a view of RL as \textit{the scientific study of agents}, a vision closely aligned with the stated goals of both RL and AI from their classic textbooks \citep{sutton2018reinforcement,russell95}, as well as cybernetics \citep{wiener2019cybernetics}. As important special cases, these agents might interact with a Markov decision process (MDP; \citeauthor{bellman1957markovian}, \citeyear{bellman1957markovian}; \citeauthor{puterman2014markov}, \citeyear{puterman2014markov}), seek to identify solutions to specific problems,  or learn in the presence of a reward signal with the goal of maximizing it, but these are not the only cases of interest.

\section{Dogma One: The Environment Spotlight}
\label{sec:dogma1_env_spotlight}

\begin{figure}[b!]
    \centering
    \includegraphics[width=0.4\columnwidth]{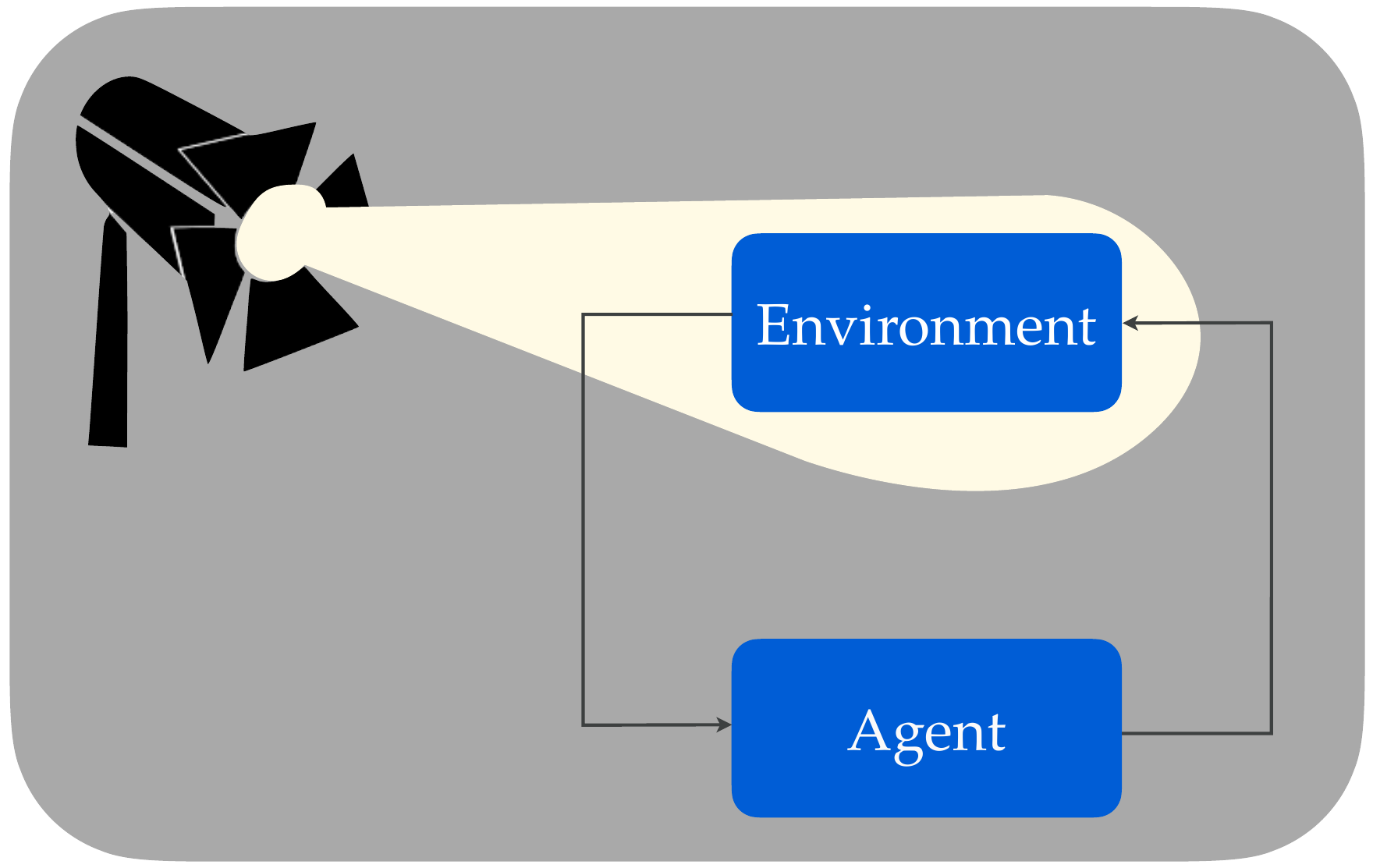}
    \caption{The first dogma, the Environment Spotlight.}
    \label{fig:d2_env_spotlight}
\end{figure}

%
The first dogma we call \textit{the environment spotlight} (\cref{fig:d2_env_spotlight}), which refers to our collective focus on modeling environments and environment-centric concepts rather than agents. For example, the agent is essentially the means to deliver a solution to an MDP, rather than a grounded model in itself.

We do not fully reject this behaviourist view, but suggest balancing it; after all the classical RL diagram features two boxes, not just one. We believe that the science of AI is ultimately about \textit{intelligent agents}, as argued by \cite{russell95}; yet, much of our thinking, as well as our mathematical models, analysis, and central results tend to orbit around solving specific problems, and \textit{not} around agents themselves. In other words, we lack a canonical formal model of an agent. This is the essence of the first dogma.

\begin{center}
\begin{minipage}{0.95\columnwidth}
\begin{mdframed}
\textsc{Dogma 1: The Environment Spotlight}
\vspace{3pt}
\hrule
\vspace{2mm}
Our collective focus on environments and environment-centric concepts, rather than agents.
\end{mdframed}
\end{minipage}
\end{center}

What do we mean when we say that we focus on environments? We suggest that it is easy to answer only one of the following two questions:
\begin{enumerate}
    \item \textit{What is at least one canonical mathematical model of an environment in reinforcement learning?}
    
    \item \textit{What is at least one canonical mathematical model of an agent in reinforcement learning?}
\end{enumerate}

%

%
The first question has a straightforward answer: the MDP, or any of its nearby variants such as a $k$-armed bandit \citep{lattimore2020bandit}, a contextual bandit \citep{langford2007epoch}, or a partially observable MDP (POMDP; \citeauthor{cassandra1994acting}, \citeyear{cassandra1994acting}). These each codify different versions of decision making problems, subject to different structural assumptions---in the case of an MDP, for instance, we make the Markov assumption by supposing there is a maintainable bundle of information we call the \textit{state} that is a sufficient statistic of the next reward and next distribution over this same bundle of information. We assume these states are defined by the environment and are directly observable by the agent at each time step for use in learning and decision making. The POMDP relaxes this assumption and instead only reveals an observation to the agent, rather than the state. By embracing the MDP, we are allowed to import a variety of fundamental results and algorithms that define much of our primary research objectives and pathways. For example, we know every MDP has at least one deterministic, optimal, stationary policy, and that dynamic programming can be used to identify this policy \citep{bellman1957markovian, blackwell1962discrete,puterman2014markov}. 
Moreover, our community has spent a great deal of effort in exploring variations of the MDP such as the Block MDP \citep{du2019provably} or Rich Observation MDP \citep{azizzadenesheli2016reinforcement}, the Object-Oriented MDP \citep{diuk2008object}, the Dec-POMDP \citep{oliehoek2016concise}, Linear MDPs \citep{todorov2006linearly}, and Factored MDPs \citep{guestrin2003efficient}, to name a few. These models each forefront different kinds of problems or structural assumptions, and have inspired a great deal of illuminating research.

%
In contrast, this second question ("what is a canonical agent model?") has no clear answer \citep{harutyunyan2020what}. We might be tempted to respond in the form of a specific kind of a popular learning algorithm, such as $Q$-learning \citep{watkins1992q}, but we suggest that this is a mistake. $Q$-learning is just one instance of the logic that \textit{could} underlie an agent, but it is not a generic abstraction of what an agent actually is, not in the same way that a MDP is a model for a broad family of sequential decision making problems. As discussed by \citet{harutyunyan2020what}, we lack a canonical model of an agent, or even a basic conceptual picture. We believe that at this stage of the field, this is becoming a limitation, and is due in part to our focus on environments.

%
Indeed, the exclusive focus on environment-centric concepts (such as the dynamics model, environment state, optimal policy, and so on) can often obscure the vital role of the agent itself. But, here we wish to reignite interest in an agent-centric paradigm that can give us the conceptual clarity we need to be able to develop and discover general principles of agency. Without such ground currently, we struggle to even precisely define and differentiate between key agent families such as "model-based" and "model-free" agents (though some precise definitions have been given by \citeauthor{strehl2006pac} \citeyear{strehl2006pac} and  \citeauthor{sun2019model}, \citeyear{sun2019model}), or study more complex questions about the agent-environment boundary \citep{jiang2019value,harutyunyan2020what}, the extended-mind \citep{clark1998extended}, embedded agency \citep{orseau2012space}, the effect of embodiment \citep{ziemke2013s,martintime},  or the impact of resource-constraints \citep{simon1955behavioral,griffiths2015rational,kumar2023continual,aronowitz2023parts} on our agents in a general way. Most agent-centric concepts are typically beyond the scope of the basic mathematical language of our field, and are consequently not featured in our experimental work.

%
\paragraph{The Alternative: Shine the Spotlight on Agents, Too.} Our suggestion is simple: it is important to define, model, and analyse agents in addition to environments. We should build toward a canonical mathematical model of an agent that can open us to the possibility of discovering general laws governing agents (if they exist), building on the work of \citet{russell1994provably}, \citet{wooldridge1995intelligent}, \citet{kenton2023discovering}, and echoing the call of \citet{sutton2022quest}. We should engage in foundational work to establish axioms that characterize important agent properties and families, as in work by \citet{sunehag2011axioms,sunehag2015rationality} and \citet{richens2024robust}. We should do this in a way that is confluent with our latest empirical data about agents, drawing from the variety of disciplines that study agents, from psychology,\footnote{Tomasello makes a similar case that the field of psychology should center around the concept of agency: "Every scientific discipline begins with a proper domain, a first principle. In biology, that proper domain or first principle is life: physical substances organized in particular ways to perform particular organismic functions. In psychology, depending on one’s theoretical predilections, that proper domain or first principle might be either behavior or mentality. But my preferred candidate would be agency, precisely because agency is the organizational framework within which both behavioral and mental processes operate." (p. 134, \citeauthor{tomasello2022evolution}, \citeyear{tomasello2022evolution}).} cognitive science, and philosophy, to biology, AI, and game theory. Doing so can expand the purview of our scientific efforts to understand and design intelligent agents.

\section{Dogma Two: Learning as Finding a Solution}
\label{sec:dogma2_learning_as_solving}

The second dogma is embedded in the way we treat the concept of learning. We tend to view learning as a finite process involving the search for---and eventual discovery of---a solution to a given task. For example, consider the classical problem of an RL agent learning to play a board game, such as Backgammon \citep{tesauro1995temporal} or Go \citep{silver2016mastering}. In each of these cases, we tend to assume a good agent is one that will play a vast number of games to learn how to play the game effectively. Then, eventually, after enough games, the agent will reach optimal play and can stop learning as the desired knowledge has been acquired.

\begin{figure*}[b]
    \centering
    \includegraphics[width=0.79\textwidth]{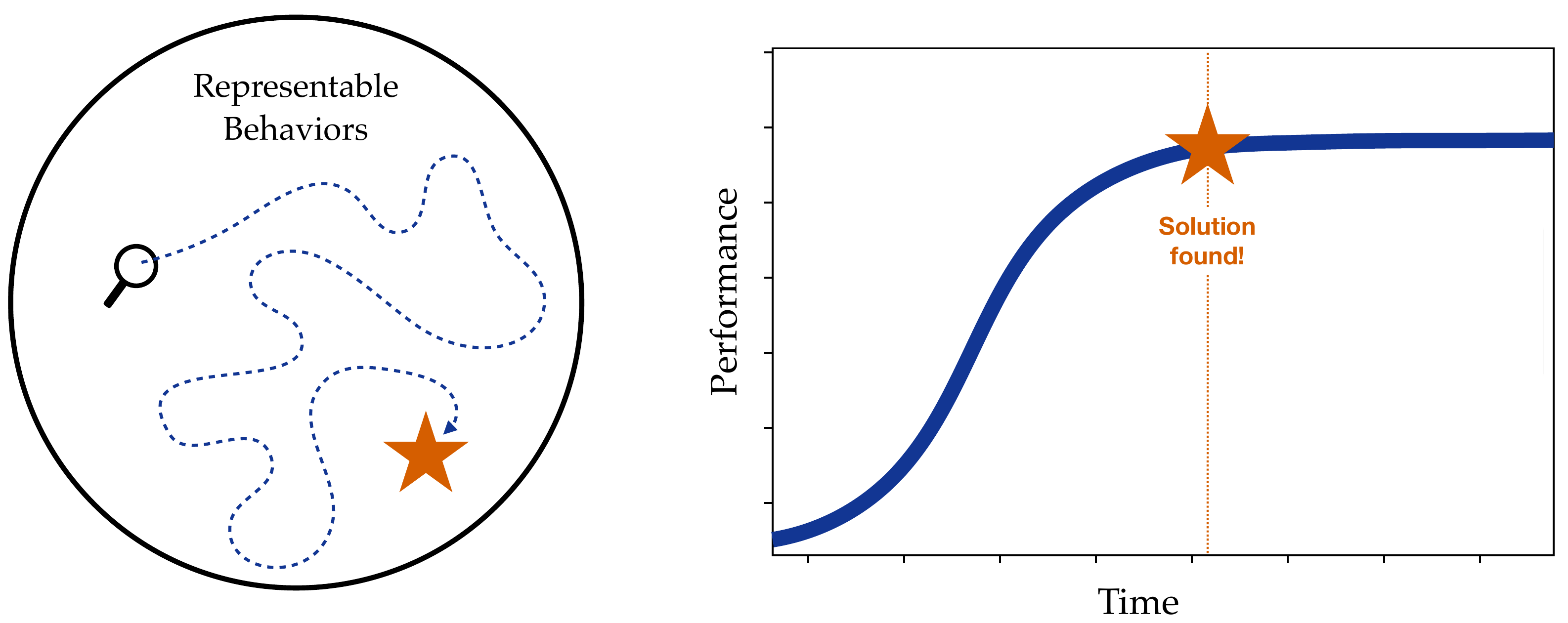}
    \caption{Dogma 2: Learning as Finding a Solution.}
    \label{fig:d2_learning_as_solving}
\end{figure*}

%
In other words, we tend to implicitly assume that the learning agents we design will eventually find a solution to the task at hand, at which point learning can cease. This is present in many of our classical benchmarks, too, such as mountain car \citep{taylor2008autonomous} or Atari \citep{bellemare2013arcade}, in which agents learn until they reach a goal. 
%
On one view, such agents can be understood as searching through a space of representable functions that captures the possible action-selection strategies available to an agent \citep{abel2023definition}, similar to the Problem Space Hypothesis \citep{newell1994unified}. And, critically, this space contains at least one function---such as the optimal policy of an MDP---that is of sufficient quality to consider the task of interested solved. Often, we are then interested in designing learning agents that are guaranteed to \textit{converge} to such an endpoint, at which point the agent can stop its search (and thus, stop its learning). This process is pictured in \cref{fig:d2_learning_as_solving}, and is summarized in the second dogma.

\begin{center}
\begin{minipage}{0.95\columnwidth}
\begin{mdframed}
\vspace{2pt}
\textsc{Dogma 2: Learning as Finding a Solution}
\vspace{3pt}
\hrule
\vspace{2mm}
Our implicit focus on designing agents that find a solution, then stop learning.
\end{mdframed}
\end{minipage}
\end{center}

%

This view is embedded into many of our objectives, and follows quite naturally from the use of the MDP as a model of the decision making problem. It is well established that every MDP has at least one optimal deterministic policy, and that such a policy can be learned or computed through dynamic programming or approximations thereof. The same tends to be true of many of the alternative learning settings we consider.

%
\paragraph{The Alternative: Learning as Adaptation.} Our suggestion is to embrace the view that learning can also be treated as adaptation \citep{barron2015embracing}. As a consequence, our focus will drift away from optimality and toward a version of the RL problem in which agents continually improve, rather than focus on agents that are trying to solve a specific problem. Of course, versions of this problem have already been explored through the lens of lifelong \citep{brunskill2014pac,schaul2018barbados}, multi-task \citep{brunskill2013sample}, and continual RL \citep{ring1994continual,ring1997child,ring2005toward,khetarpal2022towards,anand2023prediction,abel2023definition,kumar2023continual}. Indeed, this perspective is highlighted in the introduction of the textbook by \citet{sutton2018reinforcement}:
\begin{quote}
    When we say that a reinforcement learning agent’s goal is to maximize a numerical reward signal, we of course are not insisting that the agent has to actually achieve the goal of maximum reward. Trying to maximize a quantity does not mean that that quantity is ever maximized. The point is that a reinforcement learning agent is always trying to increase the amount of reward it receives. (p. 10, \citeauthor{sutton2018reinforcement}, \citeyear{sutton2018reinforcement}).
\end{quote}
This is a matter of a shift of emphasis: when we move away from optimality, how do we think about evaluation? How, precisely, can we define this form of learning, and differentiate it from others?  What are the basic algorithmic building blocks that carry out this form of learning, and how are they different from the algorithms we use today? Do our standard analysis tools such as regret and sample complexity still apply? These questions are important, and require reorienting around this alternate view of learning. We suggest that we as a community shed the second dogma and study these questions directly.

\section{Dogma Three: The Reward Hypothesis}
\label{sec:dogma3_rh}

The third dogma is the \textit{reward hypothesis} \citep{suttonwebRLhypothesis,littman2015reinforcement,christian2021alignment,abel2021reward,bowling2023settling}, which states "All of what we mean by goals and purposes can be well thought of as maximization of the expected value of the cumulative sum of a received scalar signal (reward)."

%
First, it is important to acknowledge that this hypothesis is not deserving of the title "dogma" at all. As originally stated, the reward hypothesis was intended to organize our thinking around goals and purposes, much like the expected utility hypothesis before it \citep{machina1990expected}. And, the reward hypothesis seeded the research program of RL in a way that has led to the development of many of our most celebrated results, applications, and algorithms. 

\begin{center}
\begin{minipage}{0.95\columnwidth}
\begin{mdframed}
\vspace{2pt}
\textsc{Dogma 3: The Reward Hypothesis}
\vspace{3pt}
\hrule
\vspace{2mm}
All goals can be well thought of in terms of reward maximization.
\end{mdframed}
\end{minipage}
\end{center}

However, as we continue our quest for the design of intelligent agents \citep{sutton2022quest}, it is important to recognize the nuance in the hypothesis.

%
\begin{figure*}[b!]
    \centering
    \includegraphics[width=0.7\textwidth]{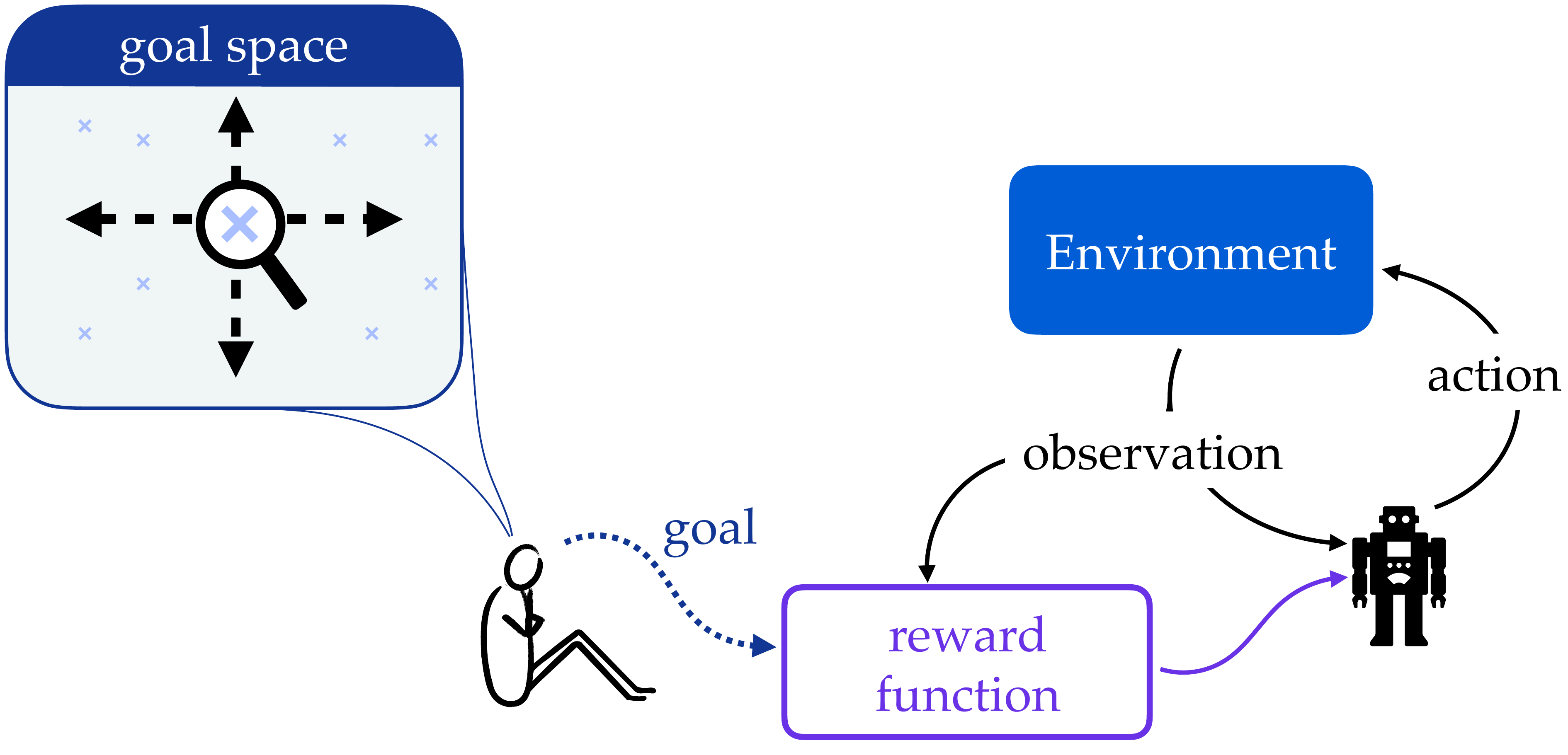}
    \caption{The third dogma, the Reward Hypothesis. Any goal that a designer might conceive of can be well thought of in terms of the maximization of a reward signal by a learning agent.}
    \label{fig:d1_rh}
\end{figure*}

In particular, recent analysis by \citet{bowling2023settling}, building on the work of \citet{pitis2019rethinking,abel2021reward} and \citet{shakerinava2022utility}, fully characterizes the implicit conditions required for the hypothesis to be true. These conditions come in two forms. First, Bowling et al. provide a pair of interpretative assumptions that clarify what it would mean for the reward hypothesis to be true or false---roughly, these amount to saying two things. First, that "goals and purposes" can be understood in terms of a preference relation on possible outcomes. Second, that a reward function captures these preferences if the ordering over agents induced by value functions matches that of the ordering induced by preference on agent outcomes. Then, under this interpretation, a Markov reward function exists to capture a preference relation if and only if the preference relation satisfies the four von Neumann-Morgenstern axioms \citep{vonneumann1953theory}, and a fifth Bowling et al. call $\gamma$-Temporal Indifference.

This is significant, as it suggests that when we write down a Markov reward function to capture a desired goal or purpose, we are \textit{forcing} our goal or purpose to adhere to the five axioms, and we must ask ourselves if it is always appropriate. As an example, consider the classical challenge on the incomparability (or incommeasurability) of values in ethics, as discussed by \citet{chang2015value}. That is, certain abstract virtues such as happiness and justice might be thought to be incomparable to one another. Or, similarly, two concrete experiences might be incommeasurable, such as a walk on the beach and eating breakfast---how might we assign measure to each of these experiences in the same "currency"? Chang notes that two items might not be comparable without further reference to a particular use, or context: "A stick can’t be greater than a billiard ball...it must be greater in some respect, such as mass or length." However, the first axiom, completeness, strictly requires that the implicit preference relation assigns a genuine preference between all pairs of experiences. As such, if we take the reward hypothesis to be true, we can only encode goals or purposes in a reward function that reject both incomparability and incommeasurability. It is worth noting that completeness in particular has been criticized by \citet{aumann1962utility} due to the demands it places on the individual holding the preference relation. Finally, the completeness axiom is not the only one restricting the space of viable goals and purposes; axiom three, independence of irrelevant alternatives, famously rejects risk-sensitive objectives as well due to the Allais paradox \citep{allais1953comportement,machina1982expected}.

\paragraph{The Alternative: Recognize and Embrace Nuance.} Our suggestion is to simply call attention to the limitations of scalar rewards, and to be open to other languages for describing an agent's goals. It is important that we are aware of the implicit restrictions we are placing on the viable goals and purposes under consideration when we represent a goal or purpose through a reward signal. We should become familiar with the requirements imposed by the five axioms, and be aware of what specifically we might be giving up when we choose to write down a reward function. On this latter point there is a profound opportunity for future work. It is also worth highlighting the fact that preferences are themselves just another language for characterizing goals---there are likely to be others, and it is important to cast a wide net in our approach to thinking about goal-seeking.

\section{Discussion}

We have here argued that the long-term vision of RL should be to provide a holistic paradigm for the science of intelligent agents. To realise this vision, we suggest that it is time to reconcile our relationship with three implicit dogmas that have shaped aspects of RL so far. These three dogmas amount to over-emphasis (1) on environments, (2) on finding solutions, and (3) on rewards as a language for describing goals. Further, we have initial suggestions on how to pursue research that makes subtle departures from these dogmas. First, we should treat agents as one of our central objects of study. Second, we must move beyond studying agents that find solutions for specific tasks, and also study agents that learn to endlessly improve from experience. Third, we should recognize the limits of embracing reward as our language for goals, and consider alternatives.


\paragraph{Open Questions.} Each of these suggestions can be translated into important research questions we encourage the community to explore further. First, \textit{what is our canonical model of an agent?} Several recent proposals have emerged, and agree on many aspects. What are the consequences of adopting one view, rather than another? Which ingredients of an agent are necessary, rather than extraneous? We suggest that it is important to think carefully about these questions, and adopt conventions for the standard model of an agent. Such a model can be used to clarify old questions, and open new lines of study around agent-centric concepts such as the agent-environment boundary \citep{todd2007mechanisms,orseau2012space,harutyunyan2020what}, embodiment \citep{ziemke2013s,martintime}, resource-constraints \citep{simon1955behavioral,ortega2011unified,braun2014,ortega2015information,griffiths2015rational,kumar2023continual,aronowitz2023parts}, and embedded agency \citep{orseau2012space}. Second, what is the goal of learning when we give up the concept of a task's solution? In other words: how do we think about learning when no optimal solution can be found? How do we begin to evaluate such agents, and measure their learning progress? Third, we suggest embracing a wide variety of views about plausible accounts of the objectives of an agent. This includes continuing to embrace classical accounts of reward maximization, but also considering varied objectives like average reward \citep{mahadevan1996average}, risk \citep{howard1972risk,mihatsch2002risk}, constraints \citep{altman2021constrained}, logical goals \citep{littman2017environment}, or even open-ended goals \citep{samvelyan2023maestro}.

\paragraph{On the term "Dogma".} The title of this paper and use of the term "dogma" are an homage to "Two Dogmas of Empiricism" by \citet{quine2000two}. The term "dogma" casts a more negative light on each of the principles than we intend (though, as \citet{kuhn1963function} notes, there is a role for dogma in the sciences). Indeed, as discussed, the reward hypothesis was originally conceived of as a \textit{hypothesis} as its name suggests. Still, it is a principle that is often taken as a presupposition that frames the rest of the field of RL similar to the way that the Church-Turing Thesis frames computation---they are both standard pre-scientific commitments that are part of most research programmes \citep{lakatos2014falsification}. The other two dogmas are both \textit{implicit} rather than conventions we regularly state openly and embrace; it is rare to see work in RL actively argue against the importance of thinking about agents or agency, for instance. Instead, it is a convention to begin most RL research by framing our research questions around dynamic programming and MDPs. In this sense, the community has been drawn to specific well-tread research paths that involve modeling environments first, rather than \textit{agents} directly. The same implicit character is true of the second dogma: due to our focus on MDPs and related models, it also tends to be the case that instances of the RL problem we study have a well structured \textit{solution} that is known to be discoverable through means such as dynamic programming or temporal difference learning. We then often use language involving an algorithm \textit{solving} a task by converging to an optimal policy, reflecting the influence of the second dogma.  It is in this sense that we take the term ``dogma" to be fitting of the first two: we tend not to question these aspects of our research programme, yet they influence much of our methods and goals.

%
It is worth noting that it is understandable why the sentiments underlying the three dogmas were adopted: by building our study from Markov models, we can make use of the suite of well-understood, efficient algorithms based on dynamic programming, thanks to the seminal work by \citet{bellman1957markovian}, \citet{sutton1988learning}, \citet{watkins1989learning}, and others. This is further supported by the way that fundamental results from stochastic approximation \citep{robbins1951stochastic} have influenced many classical results, such as the convergence of $Q$-learning by \citet{watkins1992q} or TD-learning with function approximation by \citet{tsitsiklis1996analysis}.

\paragraph{Inspiration.} We are not the first to suggest moving beyond some of these conventions.
%
The work on \textit{general} reinforcement learning by \citet{hutter2000theory,hutter2002self,hutter2004universal} and colleagues \citep{lattimore2011asymptotically,leike2016nonparametric,cohen2019strongly} has long studied RL in the most general possible setting. Indeed, the stated goal of the original work on AIXI by \citet{hutter2000theory} was "...to introduce the universal AI model" (p. 3). 
%
%
Similarly, a variety of work has explicitly focused on agents. For instance, the classical AI textbook by \citet{russell95} defines AI "as the study of agents that receive percepts from the environment and perform actions" (p. viii), and frames the book around "the concept of the intelligent agent" (p. vii).
%
\citet{russell1994provably} also feature a general take on goal-directed agents that has shaped much of the agent-centric literature that follows---the agent functions there introduced have been more recently adopted as one model of an agent \citep{abel2023convergence,abel2023definition}. \citet{sutton2022quest} proposes the "quest for a common model of the intelligent decision maker", and provides initial suggestions for how to frame this quest.
%
%
%
Work by \citet{dong2022simple} and \citet{lu2021reinforcement} have built on the traditions of agent-centric modeling, providing detailed accounts of the possible constituents of an agent's internal mechanism, similar to Sutton. Further work by \citet{kenton2023discovering} and \citet{richens2024robust} explore a causal perspective on agents, giving both concrete definitions and insightful results. Outside of AI, the subject of \textit{agency} is an important subject of discourse in its own right---we refer the reader to the work by \citet{barandiaran2009defining} and \citet{dretske1999machines} or the books by \citet{tomasello2022evolution} and \citet{dennett1989intentional} for further insights from nearby communities.
%

Similarly, a variety of work has explored alternative ways to think about goals. For instance, \citet{little2013learning} study an agent that learns a predictive model of its environment, and ground this study using the tools of information theory. This is similar in spirit to the Free-Energy Principle advocated for by \citet{friston2010free}, with recent work by \citet{hafner2020action} exploring connections to RL. Preferences have also been used as an alternative to rewards, as in preference-based RL \citep{wirth2017survey}, with a more recent line of work on RL from human feedback \citep{christiano2017deep,macglashan2016convergent,macglashan2017interactive} now playing a significant role in the current wave of language model research \citep{achiam2023gpt}. Others have proposed the use of various logical languages for grounding goals, such as linear temporal logic \citep{littman2017environment,li2017reinforcement,hammond2021multi} and nearby structures such as reward machines \citep{icarte2022reward}. Another perspective presented by \citet{shah2021benefits} explicitly contrasts the framing of assistance games \citep{hadfield2016cooperative} with reward maximization, and suggests that the former provides a more compelling path to designing assistive agents. 
Lastly, a variety of work has considered forms of goal-seeking beyond expected cumulative reward, as in ordinal dynamic programming \citep{koopmans1960stationary,sobel1975ordinal}, convex RL \citep{zahavy2021reward,mutti2022challenging,mutti2023convex}, other departures from the expectation \citep{bellemare2017distributional,bellemare2023distributional}, or by incorporating other objectives such as constraints \citep{le2019batch,altman2021constrained} or risk \citep{mihatsch2002risk,shen2014risk,wang2023near}.

%
\paragraph{Other Dogmas.} There are many other assumptions inherent to the basic philosophy of reinforcement learning that we did not discuss. For instance, it has been common to focus on agents that learn from a \textit{tabula rasa} state, rather than consider other stages of learning. We also tend to adopt the cumulative discounted reward with a geometric discounting schedule as the objective, rather than using a hyperbolic schedule \citep{fedus2019hyperbolic}, or consider the existence of environment-state rather than a partially observable setting \citep{cassandra1994acting,dong2022simple}. We take it that reflecting on these and other perspectives is also important, but that they have already received significant attention by the community.

%
\paragraph{Conclusion.} We hope this paper can reinvigorate the RL community to explore beyond our current frames. We believe this begins by embracing the vision that RL is a good candidate for a holistic paradigm of intelligent agents, and continues with a careful reflection of the values, methods, and ingredients of our scientific practice that will enable this paradigm to flourish.

\section*{Acknowledgements}

The authors are grateful Andr{\' e} Barreto and Dilip Arumugam for detailed comments on a draft of the paper, and to the anonymous RLC reviewers for their helpful feedback. We would further like to thank the many people involved in discussions that helped shape the authors' thoughts on each of the three dogmas: Sara Aronowitz, Andr{\' e} Barreto, Mike Bowling, Brian Christian, Will Dabney, Michael Dennis, Steven Hansen, Khimya Khetarpal, Michael Littman, John Martin, Doina Precup, Mark Ring, Mark Rowland, Tom Schaul, Satinder Singh, Rich Sutton, Hado van Hasselt, Ben Van Roy, and all of the members of the Agency team.

\bibliographystyle{styles/rlc}
\bibliography{main}

\end{document}